\documentclass[journal]{IEEEtran}

\usepackage{color,xcolor}
\usepackage{subfigure}
\usepackage{hyperref}
\usepackage{booktabs}
\usepackage{amssymb,mathtools,multirow}
\usepackage{amsmath,amssymb,amsfonts}
\DeclareMathOperator*{\argmax}{argmax}

\ifCLASSINFOpdf

\else

\fi

\begin{document}

\title{Unsupervised Continual Learning in Streaming Environments}


\author{Andri Ashfahani and Mahardhika~Pratama,~\IEEEmembership{Senior Member,~IEEE}
\thanks{A. Ashfahani and M. Pratama share equal contributions. Both of them are with the School of Computer Science and Engineering, Nanyang Technological University, Singapore, e-mail: andriash001@e.ntu.edu.sg; mpratama@ntu.edu.sg.}} 

\maketitle

\begin{abstract}
A deep clustering network is desired for data streams because of its aptitude in extracting natural features thus bypassing the laborious feature engineering step. While automatic construction of the deep networks in streaming environments remains an open issue, it is also hindered by the expensive labeling cost of data streams rendering the increasing demand for unsupervised approaches. This paper presents an unsupervised approach of deep clustering network construction on the fly via simultaneous deep learning and clustering termed Autonomous Deep Clustering Network (ADCN). It combines the feature extraction layer and autonomous fully connected layer in which both network width and depth are self-evolved from data streams based on the bias-variance decomposition of reconstruction loss. The self-clustering mechanism is performed in the deep embedding space of every fully connected layer while the final output is inferred via the summation of cluster prediction score. Further, a latent-based regularization is incorporated to resolve the catastrophic forgetting issue. A rigorous numerical study has shown that ADCN produces better performance compared to its counterparts while offering fully autonomous construction of ADCN structure in streaming environments with the absence of any labeled samples for model updates. To support the reproducible research initiative, codes, supplementary material, and raw results of ADCN are made available in \url{https://tinyurl.com/AutonomousDCN}.
\end{abstract}


\IEEEpeerreviewmaketitle

\section{Introduction}
\IEEEPARstart{T}{he use} of deep neural networks (DNNs) for data streams is desired because it offers the automatic features engineering step and its aptitude to handle complex problems with high-input dimension \cite{VincentDAE}. Nevertheless, such approaches are under-explored in the data stream literature due to the iterative training nature of DNNs as well as its rigid structure. The former issue rises the computational and memory issue for data streams requiring data samples to be scanned once without revisiting them again in the future \cite{GamaDataStream}. The latter issue causes the so-called retraining phase from scratch being impossible to be carried out in the resource-constrained environments of data streams and imposes relevant knowledge to be catastrophically erased by new ones \cite{li2017learning}. In addition, the fixed structure of DNNs warrants the model's capacity to be determined before the process runs - impossible to be done given the non-stationary natures of data streams. It often leads to the over-estimation problem where the network structure of DNNs are overly complex and redundant thus calling for the pruning approaches to be performed afterward \cite{Huang05ageneralized,lin2019toward}. 

{The requirement of powerful algorithms for data streams is evident in practise notably in the predictive maintenance sector. That is, data are collected from sensors continuously with a fast sampling rate \cite{pensembleplus}. This issue cannot be settled by deployment of a static model which cannot adapt to new training patterns due to non-stationary natures of a manufacturing process \cite{GamaDataStream}. Once maintenance is completed, a model is supposed to revisit its previously seen state. In other words, a model should be robust against the catastrophic forgetting problem when embracing to a new operating condition \cite{maltoni2019continuous}. That is, it must possess a knowledge retention property. Another issue is perceived in the prohibitive labeling cost because data annotation often results in a complete shutdown of the manufacturing process for visual inspection. This calls for a continual learner handling non-stationary data streams in a label-wise fashion while being robust against the issue of catastrophic forgetting.}   

The algorithmic development of DNNs for data streams has started to gain research attention. In \cite{OnlineDeepLearning}, Online Deep Learning (ODL) is proposed with the idea of hedge back-propagation where every layer has its direct connection to the output layer and produces its own local output. Despite its online training scheme, this approach still relies on the fixed structure which does not adapt to the concept drift. Incremental learning of denoising autoencoder is proposed in \cite{Zhou_incrementallearning} to grow the hidden nodes based on the loss criteria and merges redundant hidden nodes. A similar idea is shared in \cite{Huang05ageneralized} where it is built upon a radial basis function network. The above-mentioned study, however, has not been evaluated under standard data stream evaluation protocol, namely, the prequential procedure \cite{GamaDataStream}. Furthermore, these approaches still depend on the shallow network structure. It is understood that the addition of network depth is capable of boosting the capacity of a model more significantly than network width \cite{powerofdepth}. 

A fully flexible deep neural networks, namely autonomous deep learning (ADL), is proposed in \cite{ADL}. ADL features a different-depth network structure where its hidden nodes and layers are self-organized from data streams. The different-depth network structure handles the catastrophic forgetting problem in integrating a new layer where every layer produces its own output aggregated by the weighted voting scheme. This idea is extended to the context of standard multi-layer perceptron (MLP) network structure in NADINE \cite{nadine} where the adaptive memory and soft-forgetting mechanisms are designed to address the catastrophic forgetting problem during the structural evolution. Another piece of work under the roof of recurrent neural networks, namely PALM, is put forward in \cite{ferdaus2019palm}. It is created based on the teacher-forcing concept where its network structure is self-evolved to handle data streams. These approaches are hampered by their fully supervised working principle thereby imposing prohibitive labeling cost.  

One of the major bottlenecks for most data stream algorithms relies on the assumption where the true class label can be gathered shortly after data samples arrive \cite{GamaDataStream}. This fact does not coincide with the fact where at least some delay is expected by the operator to label data samples \cite{SCARGC}. It is also known that the labeling cost is varying. For instance, the fault diagnosis via manual inspection has to be performed to feed the correct class label. That is, one target class might be more difficult to gather than other classes. This issue also creates difficulty in adapting to the concept drift as the drift status has to be flagged with the absence of any target classes. Further, the aforementioned approaches are not capable of handling the catastrophic forgetting because of the presence of new tasks. That is, unable to deliver a good performance on the previously seen task after learning a new task \cite{maltoni2019continuous,sun2020continual}.

One attempt to resolve the issue of labeling cost is via the deployment of active learning by means of actively querying important samples for model updates to operators. The confidence score is proposed to actively sample important labels of data streams for model updates \cite{pensembleplus}. This method indeed reduces the labeling cost yet it often biases to under-represented classes in which the class label might be difficult to obtain. In addition, the true class label has to be immediately obtained without considering the possible delay. Another attempt to overcome the labeling cost is via the algorithmic development of a semi-supervised algorithm assuming partially labeled samples. In \cite{wang2010semi}, it makes use of the incremental learning of denoising autoencoder coupled with the hashing algorithm. A self-evolving denoising autoencoder having closed-loop configuration between the generative and discriminative learning phases is proposed \cite{devdan}. These approaches, however, do not fit to handle the infinite delay case where the true class label is only available during the warm-up phase.

The labeling cost of the data stream is addressed as the extreme latency problem where it relies only on prerecorded labels while the true class label never arrives during the process runs. COMPOSE is proposed in \cite{COMPOSE} to cope with the extreme latency problem via the computational geometry approach. In \cite{SCARGC}, SCARGC is put forward in which it adopts the pool-based approach. These two approaches are non-deep-learning approach. The problem of labeling delay is addressed via the dynamic skip connection of recurrent neural network in \cite{das2020self}. In \cite{parsnet}, the deep learning approach via the SLASH method is put forward. Our approach differs from these approaches since it is a fully unsupervised approach. That is, the training process is performed with the absence of labeled samples. The class labels are only used to induce the class-associate representations \cite{smith2019unsupervised}, i.e., {few labeled samples are only exploited to determine the class tendency of a cluster being done in the initialization phase. Note that this step is necessary because we deal with a classification task rather than a clustering task.} 

An Autonomous Deep Clustering Network (ADCN) is proposed in this paper for unsupervised data stream classification in lifelong environments. ADCN is constructed under the deep network structure consisting of a feature extractor and an autonomous fully connected layer. The feature extractor layer may adopt either the convolutional layer or MLP layer. On top of that, a flexible fully connected layer is employed. It is crafted by the deep clustering network approach with a self-evolving property. That is, both network width and depth are self-generated. Hidden nodes can be automatically generated or pruned in a flexible manner using the Network Significance (NS) method \cite{devdan} while the drift detection method is used to expand the network depth. The self-clustering mechanism is performed in the deep embedding space where every layer produces its own set of clusters and its own local output. That is, clusters are self-generated in different levels of deep embedding space to cope with any possible concept drift. A simple summation is applied to infer the final predicted output.

The overall optimization objective aims to produce the cluster-friendly latent space preventing the trivial solution \cite{liu2005toward}. In other words, it performs simultaneous feature learning and clustering under one joint optimization problem. It replaces the traditional paradigm of linear transformation having difficulty in handling complex data distribution with the deep nonlinear transformation via stacked autoencoder. It jointly optimizes both the reconstruction loss in the greedy-layer wise fashion and the clustering loss. Further, a latent-based regularization adopted from \cite{li2017learning} completes the ADCN learning policy. It is capable of resolving the catastrophic forgetting problems by creating a task-invariant network.

This paper consists of five major contributions: 1) it presents an unsupervised approach for data stream classification; 2) it proposes the deep clustering approach for data stream classification; 3) it put forward a methodology for structural learning of deep clustering networks from data streams; 4) it incorporates a latent-based regularization strategy to mitigate the catastrophic forgetting problems; 5) source codes of ADCN including all our numerical results are made public to support reproducible research. Our numerical results have demonstrated the efficacy of ADCN for handling data streams in an unsupervised manner where it produces better accuracy to those counterparts in most cases. 

The rest of this paper is structured as follows: Section 2 outlines the problem formulation; Section 3 discusses the learning policy of ADCN; Section 4 elaborates our experiments; some concluding remarks are drawn in the last section of this paper. 


\section{Problem Formulation}
Data stream problem is defined as the problem of never-ending information flow $B_1,B_2,B_3,...,B_K$ where data batch $B_1$ is sampled regularly within a specific time interval and $K$ is the number of data batches often unknown in practice. This peculiar property requires data stream $B_k$ to be handled in \textbf{a single scan} without revisiting it again in the future to assure scalable space and memory complexities. Data stream arrive with the absence of correct class labels $B_k=X_k\in\Re^{N \times u}$ where $N$ and $u$ respectively denote the size of data batch and the data dimension. ADCN handles data streams in point-by-point $N=1$ or chunk-by-chunk equally well including constant or varying chunk size. Input samples $X_k$ are paired with true class labels $Y_k\in\Re^{N\times M}$ via a particular labeling mechanism where $M$ stands for the number of target classes. Nonetheless, the labeling cost of the data stream is laborious and often calls for continual labeling efforts by operators.

The underlying objective is to build a teacher-free predictive model $f(.)$ being capable of associating an input sample $X$ to its corresponding class label $Y=f(X)$ with the absence of any true class label navigating its training process. Another important property of data streams lies in the rapidly changing data distributions affecting the joint probability distribution $P(Y|X)_t\neq P(Y|X)_{t+1}$. This issue necessitates a self-organizing property of a model where it is capable of adjusting the model's capacity in respect to distributional variations of data streams. In the realm of unsupervised data stream classification, a model has no knowledge of the desired target distribution. That is, the structural learning of a model is governed based on the change of marginal probability distribution $P(X)_t\neq P(X)_{t+1}$.

In the streaming environments, there may exist a new task where the incoming data belonging to both known or new classes come into the picture in a batch $B_k$ \cite{maltoni2019continuous}. This situation, also known as a continual learning environment, demands an algorithm to adapt to the new knowledge, without forgetting the previously seen task, namely, catastrophic forgetting \cite{sun2020continual}. One may consolidate all data and retrain the network. This practice, however, is far from the biological learning philosophy and does not fit to handle endless sequences of streaming data that are generated continuously in a rapid manner \cite{GamaDataStream}. Consequently, the predictive model $f(.)$ should be able to accumulate the knowledge, hence delivering positive transfer, as well as being able to generalize to the previous task. That is, successfully mitigating the catastrophic forgetting problem which becomes the second objective of this study.

\section{Learning Policy of ADCN}
\subsection{Network Architecture of ADCN}
The network architecture of ADCN is built upon two modules, feature extraction layer and self-evolving fully connected layer tracking any distributional changes of data streams. The feature extraction layer can be built upon either MLP or CNN layer handling both sensory data streams or visual data streams. That is, raw input signals $X$ is fed to the feature extraction layer $F(.)$ to produce natural input features $Z$. The parameters of the feature extractor are updated in the same way as in autoencoder obtaining useful features \cite{VincentDAE}. In other words, the feature extractor enables the remainder part of the network to enjoy better representation for clustering purposes. 

\begin{figure}[!t]
\centerline{\includegraphics[scale=0.52]{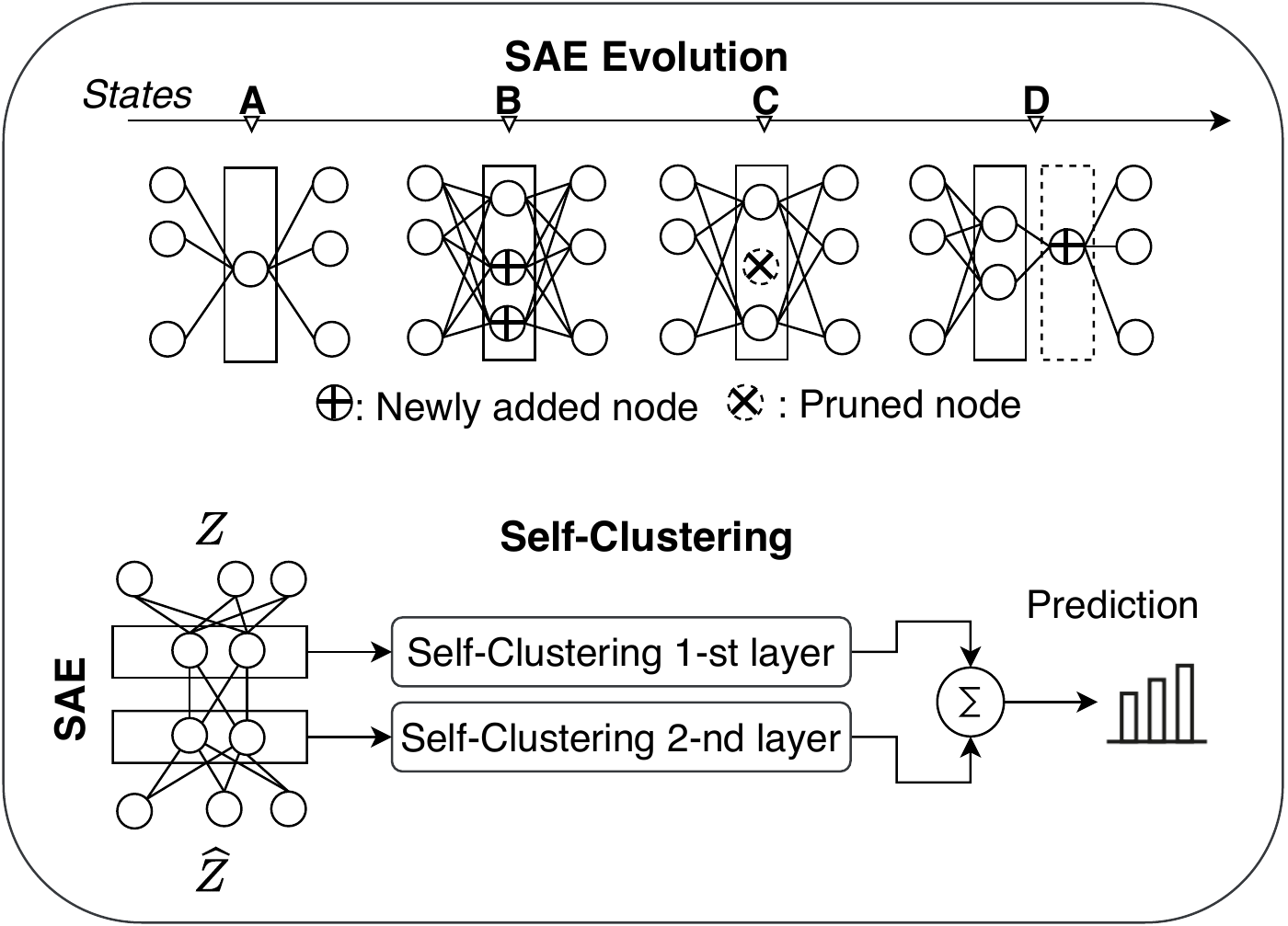}}
\caption{The ADCN network evolution. \textbf{State A}: SAE starts its learning process with a simple structure. \textbf{State B}: Two hidden nodes are added. \textbf{State C}: A hidden node is pruned. \textbf{State D}: A hidden layer is constructed. Each hidden layer is connected to a self-clustering mechanism.}
\label{fig:adcnnetworkevo}
\end{figure}

The output of the feature extractor $Z$ is injected to flexible fully connected layer which serves as Stacked Autoencoder (SAE) as illustrated in Fig. \ref{fig:adcnnetworkevo}. It projects the natural features to the lower dimensional latent space. SAE is constructed by the encoder (\ref{eq:encoder}) and decoder (\ref{eq:decoder}) trained to reconstruct the natural features as follows:
\begin{eqnarray}
   h^l = r(W^{l}h^{l-1} + b^{l}); \quad h^0 = Z \label{eq:encoder}\\
   \hat{h}^{l-1} = r((W^{l})^Th^l + d^{l}); \quad \forall l = 1,\dots, L \label{eq:decoder}
\end{eqnarray}
where $W^l\in\Re^{{R_l}\times u_l}$ and $b^l$ are respectively the SAE weight and bias in the $l-th$ layer, while $d\in\Re^{{u_l}}$ is decoder bias of the $l-th$ layer. Note that in this research a tied weight constraint is adopted. That is, the decoder weight is the transpose of the encoder weight attempting to reduce the risk of overfitting \cite{VincentDAE}. The number of hidden nodes and the number of inputs in $l-th$ layer are respectively denoted as $R_l$ and $u_l$, whereas $L$ represents the number of SAE hidden layer. A function $r(.)$ is arranged as the ReLU activation function.

The parameters of feature extractor and SAE are learned in the greedy layer-wise fashion to deal with a non-convex optimization problem. Furthermore, SAE features the self-evolving characteristic where its network width and depth are governed by the particular structural learning procedure rather than being prefixed to handle the change of marginal distributions $P(X)_t\neq P(X)_{t+1}$. In addition to the dimensionality reduction approach, the use of SAE also works to avoid the trivial solution frequently occurring in the case of a linear transformation. That is SAE functions as the nonlinear projector of natural features generating the clustering-friendly latent space \cite{liu2005toward}. 

The self-clustering mechanism is performed in every layer of SAE where its local output is produced by calculating the cluster's allegiance \cite{smith2019unsupervised} as follows: 
\begin{eqnarray}\label{allegiance}
    a^l_{c_j,h_n}=\frac{\exp{(-||c^l_{j}-h^l_{n}||)}}{\max_{j=1,...,C_l}\exp{(-||c^l_{j}-h^l_{n}||)}}
\end{eqnarray}
where $c^l_{j}$ and $C_l$ are the centroid of $j-th$ cluster and the number of clusters at $l-th$ layer of SAE, respectively. The number of cluster can be incrementally increased to encompass the uncovered representation. Until this point, it is clear that updating the ADCN parameters can be achieved without any labeled samples. The feature extractor, SAE and self-clustering mechanism can be updated in an unsupervised manner. 

When it comes to classification, several labeled samples can be provided for associating centroids with classes via Equation (\ref{allegiance}). That is, latent representations of labeled samples $h^l_{n}\in\Re^{N_m}$ are associated to the nearest centroid $c^l_{j}$. The allegiance of a cluster to the $m-th$ class is the average $a^l_{c_{j},h_{n}}$ to all labeled samples that belong to the $m-th$ class as follows:
\begin{equation}\label{allegiance2}
    A^l_{c_{j},m}=\frac{1}{N_m}\sum_{n=1}^{N_m}a^l_{c_j,h_n}
\end{equation}
where $N_m$ is the number of labeled samples belonging to the $m-th$ class. The local output of the $l-th$ layer is produced by combining the cluster's allegiance and the distance of cluster to the incoming sample $h^l$. In a nutshell, $h^l$ is classified to a class by $l-th$ layer which maximizes a combined score consisting both cluster allegiance and the distance formalized as follows:
\begin{eqnarray}
    score^l=Softmax(\sum_{j=1}^{C_l}A^l_{c_{j}}\exp{(-||c^l_{j}-h^l||)})\label{localoutput}\\
    \hat{Y}=\argmax_{m=1,...,M}\sum_{l=1}^{L} score^l_m\label{finaloutput}
\end{eqnarray}
where $score^l\in\Re^{m}$ is regarded as a score of a data point $h^l$ obtained from the $l-th$ layer indicating it tends to be predicted to which class. In addition to standardizing the score of every layer, $Softmax(.)$ is useful to ensure the partition of unity property of score. Once calculating the local score of $L$ layers SAE, the predicted label is resulted from the summation of $score^l$ across all $L$ layers as written in (\ref{finaloutput}). {Note that labeled samples are only exploited to associate a cluster to a target class as per (\ref{allegiance2}) because a classification problem is considered rather than a clustering problem here and this step can be performed in the initialization phase. Labeled samples are not used at all for model updates.}

\subsection{Structural Learning of SAE} 
\textbf{Adjustment of Network Width}: ADCN possesses a flexible network structure where its hidden nodes are automatically grown and pruned on the fly. It is governed by the network significance (NS) method \cite{devdan} derived from the bias-variance decomposition concept formalized as follows:
\begin{eqnarray}
    NS = (h^l-E[\hat{h}^l])^2+(E[(\hat{h}^l)^2]-E[\hat{h}^l]^2) \label{NS} \\
    NS = Bias(h^l)^2+Var(h^l) \nonumber
\end{eqnarray}
where $E[\hat{h}^l]$ stands for the expected reconstructed features of SAE. Note that the bias and variance expression here is based on the reconstruction loss of SAE in a greedy layer-wise manner. The growing phase is controlled by the bias term where high bias, underfitting, signifies the addition of extra hidden nodes to increase the model's capacity while high variance, overfitting, pinpoints removal of inconsequential hidden nodes to decrease the model's capacity. Assuming the normal distribution $p(h^l)$, Equation (\ref{NS}) can be solved \cite{devdan}. 

The hidden unit growing condition is formulated from the statistical process control method with dynamic confidence level as follows:
\begin{eqnarray}
    \mu_{bias}^t+\sigma_{bias}^t\geq \mu_{bias}^{min}+k_1\sigma_{bias}^{min}\rightarrow Growing \label{growing} \\
    \mu_{var}^t+\sigma_{var}^t\geq \mu_{var}^{min}+2k_2\sigma_{var}^{min}\rightarrow Pruning \label{pruning}
\end{eqnarray}
where $k_1=1.3\exp{(-Bias^2)+0.7}$ and  $k_2=1.3\exp{(-Var)+0.7}$. $k_1,k_2$ lead to dynamic confidence level in respect to the network's bias and variance. That is, new nodes are likely generated in the case of high bias whereas it does not tend to insert new nodes in the case of low bias. The same case applies in the pruning process except it reflects to the network variance instead of bias. The term $2$ is integrated in the network variance case to prevent the direct-pruning-after-adding situation. Finally, $\mu_{bias}^{min},\sigma_{bias}^{min},\mu_{var}^{min},\sigma_{var}^{min}$ are reset once (\ref{growing}), (\ref{pruning}) are satisfied.

\textbf{Adjustment of Network Depth}: ADCN adopts an elastic network width which can be expanded to cope with concept drift. The adjustment of network depth plays a vital role to improve network generalization power because it substantiates the network's capacity \cite{powerofdepth}. A drift detection method is applied here to govern the addition of network depth and is based on Hoeffding's bound statistic. Note that the point of interest here is in the change of marginal distribution, virtual drift, rather than the conditional distribution, real drift, because of the unsupervised nature of ADCN. The drift detection method initiates by finding the cutting point, a point where the population means increases. A cutting point, $cut$, is identified if the following condition comes across: 
\begin{equation}
    \hat{S}+\epsilon_{S}\leq\hat{T}+\epsilon_{T}
\end{equation}
where $S\in\Re^{2N}$ is the extracted features $Z$ produced using previous and current batch data $[B_{k-1};B_{k}]$, whereas $T\in\Re^{cut}$ and $cut\leq 2N$. Two batches of data are exploited in this strategy aiming to increase the sensitivity of the drift detector in identifying a drift on the stable latent representation. {Note that the previous data batch $B_{k-1}$ is only used to induce the drift detection method and not exploited anywhere else. It is discarded once used}. The hypothetical cutting point is arranged as $[25\%,50\%,75\%]\times 2N$ to overcome the false alarm while $\hat{S},\hat{T}$ stands for the statistics of $S,T$, respectively. On the other hand, the error bound $\epsilon_{S,T}$ are defined from Hoeffding's bound as follows:
\begin{equation}
    \epsilon_{S,T}=\sqrt{\frac{1}{2\times size}\ln\frac{1}{\alpha_x}}
\end{equation}
where $size$ denotes the cardinality of the data matrix $S,T$, while $\alpha_x$ is the significance level. Note that the significance level is inversely proportional to the confidence level of $1-\alpha_x$. 

Once finding the cutting point $cut$, the drift condition can be signalled if $|\hat{S}-\hat{T}|\geq\epsilon_{d}$. There exists another condition, warning, where a drift still requires further investigation with the next data batch. The warning condition, $\epsilon_{w}\leq|\hat{S}-\hat{T}|\leq\epsilon_{d}$, is akin to the drift condition unless a higher significance level is utilized $\alpha_{d}<\alpha_{w}$ leading to a lower confidence level. The error bound drift and warning $\epsilon_{d},\epsilon_{w}$ are written as follows:
\begin{equation}
    \epsilon_{d,w}=(b - a)\times\sqrt{\frac{size-cut}{2\times cut\times size}\ln\frac{1}{\alpha_{d,w}}}
\end{equation}
where $[a,b]$ represents the interval of $S$. The concept drift induces the introduction of a new layer {where the number of nodes of a new layer is set as the half of the number of nodes of a previous layer. This strategy is designed to induce the nonlinear feature reduction mechanism and to prevent an over-complete network.} Initial unlabeled data is exploited to construct a new layer if the next data batch returns a drift condition. Conversely, a stable condition only leads to the parameter learning phase. It is worth mentioning that the catastrophic forgetting problem is not incurred here since the greedy layer-wise training scenario is implemented. Moreover, the self-clustering mechanism is implemented in every layer thereby generating their own local output. The overall network evolution of ADCN is illustrated in Fig. \ref{fig:adcnnetworkevo}. 

\textbf{Self-Clustering Mechanism}: ADCN applies the self-clustering mechanism in the deep embedding space - every layer of SAE which aims to find the clustering-friendly latent space thus resulting in the high generalization power. The self-clustering mechanism is implemented here where clusters are self-evolved from data streams to deal with variational distributions. Suppose that $D(x,y)$ denotes the L2 distance between two variables $x,y$, a new cluster is added if the following condition is fulfilled: 
\begin{eqnarray}
    D(h^l,c^l_{win})>\mu^l_{win}+k^l_{win}\sigma^l_{win} \label{clusteradd}\\
    k^l_{win}=2\exp{(-D(h^l,c^l_{win})})+2 \nonumber
\end{eqnarray}
where $c^l_{win}$ is the winning cluster of the $l-th$ latent space, while $\mu^l_{win},\sigma^l_{win}$ are the mean and standard deviation of $D(h^l,c^l_{win})$, respectively. The winning cluster is that having the most adjacent distance to the cluster $win\rightarrow \min_{i=1,...,C_l}D(h^l,c^l_{win})$. The dynamic constant $k^l_{win}$ is employed to realize a dynamic confidence degree. The condition in (\ref{clusteradd}) indicates that a new cluster tends to be appended if the winning cluster is incapable of covering the current latent representation $h^l$. It also signifies the presence of concept drift since a data point is not properly covered by any cluster. A new cluster is parameterized as $c_{C_l+1}^l=h^l$ and $Sup_{C_l+1}^l=1$ where those respectively represent the centroid and support of the new cluster.

\subsection{Parameter Learning Strategy}
\textbf{Simultaneous Feature Learning and Clustering}:
it is performed via joint cost function taking into account both reconstruction loss and the clustering loss. The reconstruction loss avoids the trap of trivial solutions by means of the nonlinear dimensional reduction via SAE mapping the latent space back to the extracted feature space. The use of clustering loss aims to discover the clustering-friendly latent space. The joint optimization problem $L_U$ is expressed as follows:
\begin{equation}
    L_U=L(X,\hat{X}) + \sum_{l=1}^{L}(L(h^l,\hat{h}^l)+\frac{\alpha}{2}||h^l-c^l_{win}||_2)\label{loss}
\end{equation}
where $L_U$ is the unsupervised learning loss and $L(.)$ represents the mean squared error (MSE) loss. The clustering loss strength $\alpha$ is a constant controlling the amount of update from the clustering loss. In other words, it finds a balance between the clustering-friendly latent space and latent space best representing the original feature space. The first and second terms are designed to update the feature extractor and to learn the latent structure of data space via $L$ multiple nonlinear mapping of SAE. Note that it also functions as the nonlinear dimensionality reduction addressing the bottleneck of simple linear mapping, namely the trivial solution. The third part of Equation (\ref{loss}) is obtained from the K-means loss function simultaneously solved with the reconstruction loss in the greedy-layer wise fashion. It is worth noting that the clustering mechanism takes place in every layer of SAE as depicted in Fig. \ref{fig:adcnnetworkevo} where ADCN features an expandable network depth controlled by the drift detection mechanism. Note that our approach differs from \cite{liu2005toward} since their clustering process is only localized into the bottleneck layer.

Simultaneous feature learning and clustering is performed in the greedy layer-wise fashion and addressed via alternate optimization between SAE's parameters and the cluster's parameters. That is, the cluster's parameters are fixed when updating SAE's parameters. SAE's parameters are optimized via the stochastic gradient descent (SGD) approach. Also, it is understood that parameter initialization is one of the important parts of DNN training \cite{VincentDAE}. Because of this reason, a small pre-training phase is executed before the streaming process run to initialize the network parameters. As many as $N_{init}$ samples are exploited in this phase. The streaming process is started after $nE$ epochs are achieved. This process is also carried out whenever there is a new layer or at the beginning of a new task. Note that the learning process in the streaming phase is conducted in a single-pass learning fashion as an effort to cope with the rapid nature of the data stream.



\textbf{Centroid Update}:
the cluster's centroid is updated in the winner-takes-all fashion if (\ref{clusteradd}) is violated. That is, only the winning cluster having the closest distance to a data sample is adjusted. The centroid's update is formulated as follows:
\begin{equation}\label{clusterupdate}
    c_{win}^{l}=c_{win}^{l}-\frac{(c_{win}^l-h^l)}{Sup_{win}^l+1};\quad Sup_{win}^{l}=Sup_{win}^{l}+1
\end{equation}
where $c_{win}^l,Sup_{win}^{l}$ are the centroid and cardinality of the winning cluster, respectively. Equation (\ref{clusterupdate}) adjusts more aggressively for a cluster with low number of population whereas it updates more grecefully for a cluster with high number of population. Furthermore, a data sample $h^l$ is assigned to the winning cluster increasing the cardinality of the winning cluster. Equation (\ref{clusterupdate}) is also seen as the SGD method where the learning rate is varied as the cluster's population.

One major cause of trivial solutions is that all data are assigned to a single cluster. To further avoid this issue, a simple trick from feature quantization \cite{johnson2019billion} is adopted here. That is, the empty clusters are automatically reassigned during the self-clustering process as suggested by \cite{Caron_2018_ECCV}. Practically, a high populated cluster is randomly selected whenever there is an empty cluster. We then use the selected cluster centroid with a small random perturbation as the new centroid for the empty cluster.

\textbf{Solution of Catastrophic Forgetting}:
up to this point, ADCN learning policy is able to handle unsupervised learning problem in streaming environments. Nonetheless, it is still prone to catastrophic forgetting since there is no mechanism to constraint the parameter movement while learning a new task. Here, a latent-based regularization $L_{CL}$ adopted from \cite{li2017learning} is added to Equation (\ref{loss}) for resolving the issue. The joint optimization problem for solving unsupervised continual learning $L_{UCL}$ can be written as follows:
\begin{eqnarray}
    L_{UCL}= L_U + \sum_{iT = 1}^{nT} \lambda_{iT} L_{BCE}(\hat{X},\hat{X}_{CL}^{iT})\label{eq:latentregularization}\\
    \lambda_{iT} = \begin{cases}
0, & nT = 1\\
\beta (1 - \frac{M_{iT}}{\sum_{jT\leq iT} M_{jT}}), & nT > 1
\end{cases}
\end{eqnarray}
where $nT$ is the number of tasks. The second term of Equation (\ref{eq:latentregularization}), namely $L_{CL}$, mitigates the catastrophic forgetting by enforcing the reconstructed output $\hat{X}$ stability. That is, it forces the reconstructed output of a network in the current task $\hat{X}$ to stay close to the reconstructed output of networks in all previous tasks, $\hat{X}_{CL}^{iT}, \forall iT = 1,\dots,nT$, given the current task input $X$ and thus creating a task-invariant network. In this research, binary cross-entropy loss $L_{BCE}$ is employed instead of knowledge distillation loss attempting to simplify the implementation in streaming environments. It is
achievable without loss of generalization as explained in \cite{li2017learning}.



The regularization strength $\lambda_{iT}\in[0,\beta]$ is designed to incrementally decrease as the number of task increases \cite{maltoni2019continuous}, where $M_{iT}$ denotes the number of classes in the current task. This is reasonable since the importance of the previous task should increase with the total number of classes learned. Here, it is also modified by introducing a constant $\beta$ increasing the amount of regularization as an effort to compensate for the single-pass nature of ADCN. Whenever there are more than one task ($nT > 1$) observed by ADCN, the latent-based regularization $L_{CL}$ is added to (\ref{loss}) to update the network parameters. The optimization of $L_{CL}$ is conducted in an end-to-end fashion since it targets all network parameters.

The regularization (\ref{eq:latentregularization}) is inspired by the proposed method in \cite{li2017learning} to overcome the catastrophic forgetting problem in a supervised learning situation. Here, it is extended as the latent-based regularization forcing the reconstructed output of the current task to behave similarly to the output of all previous tasks. Further, it is executed in an unsupervised manner.










\section{Proof of Concepts}
This section demonstrates the classification performance of ADCN on benchmark datasets. The performance of ADCN was simulated on two scenarios: standard unsupervised learning and unsupervised learning in a continual learning environment. The ADCN learning policy together with those two simulation scenarios are presented in Algorithm 1, and Fig. S-1 and S-2 of the supplemental document. There are also provided a list of symbols and acronyms to facilitate the reader in Table S-I and S-II.

\subsection{Unsupervised Learning Problem}
\textbf{Experimental Setup}:
here, all consolidated algorithms are required to classify streaming unlabeled samples where there is only one task $nT=1$. The data arrive batch-by-batch where the batch size is $N=1000$. Since the prequential procedure is employed here, the accuracy is calculated per batch and the final result is the average accuracy of all batches, namely prequential accuracy (preq. acc.). It reflects the performance of an algorithm in handling concept changes \cite{GamaDataStream}.

In the pre-training phase, we simply consider $N_{init} = \{1K,\:5K\}$ and $nE = 50$. At the end of the first batch, a limited amount of labeled data per class ($N_m=500$) is introduced to evaluate the prequential accuracy. The labeled data remain unchanged until the streaming phase is completed. This simulates a real-world scenario where the operators only release the labels once and let the algorithms autonomously adapt to any incoming concept changes \cite{SCARGC}. The reported results are the average over 5 runs. In each run, the same unlabeled data stream is handled. In other words, it tests the consistency of ADCN in dealing with a given problem. These protocols are also applied to all baselines.

\textbf{Benchmark Datasets}:
we use synthetic and real-world datasets to demonstrate the efficacy of ADCN. Four of them are in the form of unstructured data, i.e., MNIST \cite{mnist}, Kuzushiji-MNIST (KMNIST) \cite{clanuwat2018deep},  Fashion-MNIST (FMNIST) \cite{xiao2017fashion} and CIFAR10 \cite{schwarz2018progress}. To induce the covariate drift to those data, the image data are rotated to arbitrary angles in the range of $[a,b]$. That is, the images are rotated 4 times $[0,5] \longrightarrow [6,10] \longrightarrow [11,15] \longrightarrow [0,15]$ which leads to two drift types: abrupt and recurring. ADCN performance is also tested on the structured data, i.e., SEA \cite{SEA}, Hyperplane \cite{MOA} and Credit Card Default \cite{creditcard} datasets. In addition to testing ADCN performance in handling non-stationary data, the use of structured data also tests the generalizability of the proposed framework, since instead of CNN, the MLP network is employed as the feature extractor. The datasets properties are summarized in Table \ref{datasetsproperties}.
\begin{table}[t!]
\caption{Properties of the Dataset.}
\label{datasetsproperties}
\begin{center}
\scalebox{0.9}{
\begin{tabular}{lrrrrr}
\toprule 
Datasets & $u$ & \#C & \#Samples & $N_{init}$ & Char.\tabularnewline
\midrule 
MNIST and KMNIST & $28\times28$ & 10 & 70K & 5K & R, U\tabularnewline
FMNIST & $28\times28$ & 10 & 70K & 5K & Syn, U\tabularnewline
CIFAR10 & $32\times32\times3$ & 10 & 60K & 5K & R, U\tabularnewline
SEA & 3 & 2 & 100K & 1K & Syn, S\tabularnewline
Hyperplane & 4 & 2 & 120K & 1K & Syn, S\tabularnewline
Credit Card Default & 24 & 2 & 30K & 1K & R, S\tabularnewline
\bottomrule
\end{tabular}
}
\end{center}
\footnotesize{Characteristics of the data (Char.): real-world data (R), synthetic data (syn), unstructured data (U), structured data (S). }
\end{table}

\textbf{Algorithms and Parameters}:
together with ADCN, we also trained Deep Clustering Network (DCN) \cite{liu2005toward}, Autoencoder followed by K-means (AE+KMEANS) and SCARGC \cite{SCARGC}. DCN is a two-stage approach which optimizes its network and clusters via joint optimization problem in (\ref{loss}) yet it does not incorporate evolving mechanism. AE+KMEANS is similar to DCN yet it optimizes its network and clusters alternately. SCARGC \cite{SCARGC} is also considered as the baseline. It is a non-DNN method specifically designed for the infinite delay problem and considered as a state-of-the-art algorithm. Two versions of SCARGC built upon SVM (SCARGC+SVM) and KNN (SCARGC+KNN) classifiers are utilized here.

On MNIST-based problems and CIFAR10 dataset, ADCN employs CNN as the feature extractor. The encoder part consists of two convolutional layers with 16 and 4 filters and max-pooling layers in between, whereas the decoder part is constructed by two transposed convolution operators with 4 and 16 filters. On structured datasets, ADCN put forward a two-layered MLP as a feature extractor where the number of nodes in each layer is set as $4\times u$. A tied weight constraint is applied to construct the decoder part. The initial node of SAE $R_1$ is simply set as 96 and $2\times u$ for unstructured and structured data cases, respectively. Finally, the ReLU activation function is utilized to induce nonlinear transformation in the intermediate layers, whereas the decoder output applies the Sigmoid activation function obtaining the normalized reconstructed input. 

The learning rate and momentum coefficient of SGD are set as $0.01$ and $0.95$. The weight decay strength is set at $5 \times 10^{-5}$ attempting to avoid overfitting. ADCN parameters $\{\alpha_{x},\alpha_{d},\alpha_{w}\}$ and $\alpha$ are set to $\{0.001,0.001,0.005\}$ and 0.01 which controls the drift rate and the amount of update from clustering loss. Note that all of these hyperparameters remain unchanged in all experiments to demonstrate the non-ad-hoc nature of ADCN. The learning batch size is selected as 16, whereas on CIFAR10, it is simply set as 128. To certify a fair comparison, all baselines are re-implemented in the same simulation scenario and procedure. For DNN-based methods, the hyperparameters and network structures are the same as those applied in ADCN, whereas the hyperparameters of non-DNN methods are re-tuned. It aims to obtain better performance, thereby providing a more competitive experimental setting for testing out ADCN. 

\textbf{Results}: 
from Table \ref{unsupervisedlearningresults}, ADCN delivered up to $76\%$ performance improvement over consolidated baselines in terms of accuracy. It achieved the best predictive performance on 5 of 6 problems. Further, the t-test confirmed that this result is statistically significant (P $<$ 0.05) in almost all cases. In addition to demonstrating the efficacy of ADCN learning policy, this also signifies that the evolving mechanisms of ADCN can generate appropriate network and cluster complexity for a given unsupervised data stream problem. ADCN predictive performance was comparable to DCN and AE+KMEANS on the SEA and Hyperplane problems. The number of initial clusters is believed to become the main cause. Note that the number of clusters of those two methods was 100 from the beginning of the training process, whereas ADCN started the process with only 2 clusters. As a result, those baselines could accommodate more knowledge enabling to deliver better predictive performance.
\begin{table}[t!]
\caption{Performance Metrics and Model Complexity on Unsupervised Learning Scenario.}
\label{unsupervisedlearningresults}
\begin{center}
\scalebox{0.8}{
\begin{tabular}{llccccr}
\toprule
Dataset & Methods & NoC & NoF & Depth &  Preq. Acc. (\%) \\
\midrule
MNIST & ADCN & 2277 $\pm$ 51 & 117 $\pm$ 17 & 1 & \textbf{86.49 $\pm$ 0.56} \\
& DCN & 500 & 96 & 1 & 80.38 $\pm$ 0.59$^\times$ \\
& AE+KMEANS & 500 & 96 & 1 &  80.11 $\pm$ 0.75$^\times$ \\
& SCARGC+SVM & 500 & 784 & N/A &  70.24 $\pm$ 1.41$^\times$ \\
& SCARGC+KNN & 500 & 784 & N/A & 9.89 $\pm$ 0.07$^\times$\\
\midrule
KMNIST & ADCN & 2178 $\pm$ 35 & 128 $\pm$ 23 & 1 & \textbf{82.48 $\pm$ 0.65} \\
& DCN & 500 & 96 & 1 & 76.32 $\pm$ 1.14$^\times$ \\
& AE+KMEANS & 500 & 96 & 1 &  76.48 $\pm$ 1.26$^\times$ \\
& SCARGC+SVM & 500 & 784 & N/A &  59.46 $\pm$ 0.86$^\times$ \\
& SCARGC+KNN & 500 & 784 & N/A & 10.03 $\pm$ 0.06$^\times$\\
\midrule
FMNIST & ADCN & 2533 $\pm$ 72 & 107 $\pm$ 4.35 & 1 & \textbf{73.04 $\pm$ 1.17} \\
& DCN & 500 & 96 & 1 & 70.23 $\pm$ 0.72$^\times$ \\
& AE+KMEANS & 500 & 96 & 1 &  69.09 $\pm$ 0.61$^\times$ \\
& SCARGC+SVM & 500 & 784 & N/A &  62.79 $\pm$ 2.11$^\times$ \\
& SCARGC+KNN & 500 & 784 & N/A & 10.10 $\pm$ 0.08$^\times$\\
\midrule
CIFAR10 & ADCN & 2451 $\pm$ 66 & 101 $\pm$ 4 & 1 & \textbf{27.80 $\pm$ 0.29} \\
& DCN & 500 & 256 & 1 & 27.05 $\pm$ 0.20$^\times$\\
& AE+KMEANS & 500 & 256 & 1 &  26.94 $\pm$ 0.27$^\times$\\
& SCARGC+SVM & 500 & 1025 & N/A & 12.90 $\pm$ 1.00$^\times$ \\
& SCARGC+KNN & 500 & 1025 & N/A & 10.09 $\pm$ 0.08$^\times$\\
\midrule
SEA & ADCN & 51 $\pm$ 6 & 7 $\pm$ 2 & 1 & \textbf{86.16 $\pm$ 1.99} \\
& DCN & 100 & 6 & 1 & 85.44 $\pm$ 0.96 \\
& AE+KMEANS & 100 & 6 & 1 &  85.43 $\pm$ 0.74 \\
& SCARGC+SVM & 500 & 3 & N/A &  82.42 $\pm$ 5.86$^\times$ \\
& SCARGC+KNN & 500 & 3 & N/A & 57.89 $\pm$ 4.34$^\times$ \\
\midrule
Hyperplane & ADCN & 98.80 $\pm$ 15 & 8 $\pm$ 1 & 1 & 83.88 $\pm$ 2.22 \\
& DCN & 100 & 8 & 1 & 83.96	$\pm$ 0.75 \\
& AE+KMEANS & 100 & 8 & 1 &  \textbf{84.12 $\pm$ 0.87} \\
& SCARGC+SVM & 500 & 4 & N/A &  78.15 $\pm$ 3.74$^\times$ \\
& SCARGC+KNN & 500 & 4 & N/A & 50.11 $\pm$ 0.05$^\times$\\
\midrule
Credit Card & ADCN & 501 $\pm$ 41 & 110 $\pm$ 3 & 6 & \textbf{74.33 $\pm$ 3.99} \\
Default & DCN & 100 & 48 & 1 & 64.17 $\pm$ 1.07$^\times$ \\
& AE+KMEANS & 100 & 48 & 1 &  61.23 $\pm$ 1.24$^\times$ \\
& SCARGC+SVM & 500 & 24 & N/A &  54.38 $\pm$ 7.64$^\times$ \\
& SCARGC+KNN & 500 & 24 & N/A & 44.97 $\pm$ 5.28$^\times$\\
\bottomrule
\end{tabular}
}



\end{center}
\footnotesize{NoC: Number of clusters, NoF: Number of latent features for clustering,\\ $^{\times}$: Indicates that the numerical results of the respected baseline and ADCN are significantly different.}
\end{table}

One can see that ADCN outperforms SCARGC by a large margin in all cases in terms of accuracy. This demonstrates the advantage of utilizing the self-clustering mechanism in deep embedding space. We notice that the performance of SCARGC in these experiments was unsatisfactory, probably because it performs a direct input-space clustering method which may experience a saturating performance on high input dimension datasets since the Euclidean distance is ineffective to deal with high dimension problems \cite{aggarwal2001surprising}. Further, the direct input-space clustering is less likely to obtain a better solution since the characteristic of input features are usually less stable. Note that the clustering works better on top of stable features \cite{Caron_2018_ECCV}.

Another important finding is also shown in Table \ref{unsupervisedlearningresults} where ADCN predictive performance is better than DCN and AE+KMEANS in almost all cases. This finding highlights the benefit of the evolving mechanism embraced by ADCN. The autonomous SAE is capable of increasing or decreasing the network capacity on demand, whereas the self-clustering evolution aims to accommodate uncovered latent space representations crafting a new knowledge. The comparison against AE+KMEANS also suggests that the combination of structural evolution and joint optimization in (\ref{loss}) is able to work hand-in-hand to find a solution for a given problem.

The performance and network evolution of ADCN is pictorially exhibited in Fig. S-3 of supplemental document illustrating the Credit Card Default problem. The flexible nature of ADCN is demonstrated where the growing and pruning mechanisms are implemented. Furthermore, the evolution of accuracy is shown to rapidly increase and stable over time highlighting the efficacy of ADCN learning policy. The accuracy decreases in the case of drift but timely responded with the introduction of new nodes and new layers. From the figure, the evolving nature of the self-clustering mechanism is also demonstrated where the number of clusters can be evolved whenever there are uncovered latent features.

\subsection{Unsupervised Continual Learning Problem}
\textbf{Experimental Setup}: ADCN capability in handling unsupervised continual learning problems is tested here where the number of tasks is greater than one ($nT>1$). We consider two continual learning scenarios in creating different tasks, i.e., new instances (NI) and new classes (NC) \cite{maltoni2019continuous}. The former scenario is defined as a situation where the newly seen data of the same classes come into the picture in the following batches possessing different distribution, whereas the latter scenario defines a case where the newly seen data of different classes arrive in subsequent batches. 

Besides prequential accuracy, we also measure task accuracy (task acc.), positive backward transfer (BWT) and positive forward transfer (FWT) as suggested in \cite{gem2017paz}. In a nutshell, those three metrics are commonly used in continual learning, and measure an algorithm's capability in both handling catastrophic forgetting and utilizing the past knowledge to improve its prediction in the incoming task. Using them together satisfies to determine the efficacy of a continual learning algorithm. Note that BWT and FWT is a value within $-\infty$ to $+\infty$, with a high positive value being the best performance in mitigating catastrophic forgetting problems and in exhibiting positive knowledge transfer.

Unless otherwise stated, a similar setup as in the first simulation scenario is adopted. In each task, all consolidated algorithms only have access to $(N_{init}/nT)$ initial unlabeled data of the current task for pre-training purpose. After that, all algorithms process streaming unlabeled samples batch-by-batch in a one-pass learning fashion. Several labeled samples per class $(N_m/nT)$ are incrementally provided according to the task for performing classification. We simply consider $[N_{init},N_m]=[5000,500]$. In each task, we also let a batch of data remain unseen for calculating task accuracy, BWT and FWT. That is, an algorithm is required to perform classification on all previously seen tasks at the end of every task. The simulation scenario of unsupervised continual learning is illustrated in Fig. S-2 of the supplemental document.

\textbf{Benchmark Datasets}: Rotated MNIST (RMNIST) \cite{gem2017paz}, Permuted MNIST (PMNIST) \cite{kirkpatrick2017overcoming}, Split MNIST (SMNIST) \cite{DeepExpandable} are considered as benchmarks here which are created from the real-world handwritten digit dataset MNIST \cite{mnist}. {In addition to the three-digit recognition problems, the Split CIFAR10 (SCIFAR10) problem \cite{schwarz2018progress} is included in our experiment. CIFAR10 is a popular dataset used to evaluate the performance of an image classification algorithm. It consists of 50K training images and 10K testing images incorporating 10 classes: airplane, automobile, bird, cat, deer, dog, frog, horse, ship and truck \cite{krizhevsky2009learning}.  Besides the classes are completely mutually exclusive \cite{krizhevsky2009learning}, it has been demonstrated in \cite{chan2018t} that the nearest neighbour cluster of CIFAR10 under the L2 distance metric is more unclear than that of MNIST. This causes CIFAR10 more difficult to classify than MNIST, thereby providing a more competitive benchmark for testing out ADCN.}


RMNIST and PMNIST problems are respectively generated via image rotation and image pixel permutation realizing the NI scenario. Four tasks are considered here ($nT=4$) generated by applying four sets of rotation angle, i.e., $\{[0,30], [31,60], [61,90], [91,120]\}$, and four random permutations to images. SMNIST and SCIFAR10 are generated by splitting the MNIST dataset in sequence based on its classes (0/1, 2/3, 4/5, 6/7, 8/9) creating five tasks ($nT=5$). SMNIST and SCIFAR10 problems realize the NC scenario. Finally, we can have 15K samples (for RMNIST and PMNIST), 13K samples (for SMNIST) and 11K samples (for SCIFAR10) as streaming unlabeled data in each task.

\textbf{Algorithms and Parameters}: we compare the proposed ADCN with DCN, AE+KMEANS and {Self-Taught Associative Memory (STAM)} \cite{smith2019unsupervised}. To make their learning policy fit in a continual learning environment, we arm DCN and AE+KMEANS with the state-of-the-art methods in the field, i.e., learning without forgetting (LwF) \cite{li2017learning} and synaptic intelligence (SI) \cite{SI}. In a nutshell, LwF and SI are capable of mitigating catastrophic forgetting problem respectively via controlling output stability and preventing the important parameters of the previous task to move away from the optimal solution. Note that they are originally operated in supervised learning. Here, those methods are used to update the autoencoder network in an unsupervised manner. {On the other hand, STAM is an unsupervised continual learning algorithm consisting of a hierarchy of increasing receptive field, online clustering, novelty detection and a dual-memory that stores the prototype of data \cite{smith2019unsupervised}. It works in the similar manner as ADCN, with the exception that STAM uses a non-neural network model as the feature extractor.}

On RMNIST, SMNIST and {SCIFAR10} problems, all consolidated algorithms utilize the same network structure as that on the MNIST and {CIFAR10} problems in the first experiment scenario. A two-layered MLP is put forward as a feature extractor instead of CNN to deal with the PMNIST problem with $[1000,500]$ hidden nodes in each layer. It is reasonable since the PMNIST problem requires a network to take into account all image pixels without any exception which is unachievable by CNN. Unless otherwise noted, the ADCN and SGD hyperparameters in the second experiment adopt the same value as in the first simulation scenario. The maximum regularization strength is set as $\beta=5$ which is also applied in LwF. The regularization strength of SI is set as 0.2 in the first task and 0.8 in the remaining tasks. These values are fixed during experiments. In the case of SCIFAR10, the learning process in streaming phase applies a small epoch of 10 times which is still applicable in streaming environments. 

\textbf{Results}: 
it is reported in Tables \ref{unsupervisedCL} that ADCN produces the highest task and prequential accuracy in RMNIST and SCIFAR10 problems. The t-test confirms that these results are statistically significant (P $<$ 0.05). The rejection of the null hypothesis indicates that ADCN's predictive accuracy is significantly better than other baselines. In terms of resolving the catastrophic forgetting problem, ADCN delivered a -13-point on average in BWT. This result is achievable since ADCN is capable of maintaining its output stability and storing the previous knowledge via latent-based regularization and self-clustering mechanism. Interestingly, the stored knowledge also helps to improve the predictive performance in incoming tasks indicated by an FWT of 18 on average, which is the highest value among the consolidated algorithms. This signifies that the ADCN learning policy is also suitable for addressing a transfer learning problem where the main objective is to obtain better predictive performance only in the target task after learning a source task.
\begin{table}[t!]
\caption{Performance Metrics on Continual Learning Scenario.}
\label{unsupervisedCL}
\begin{center}
\scalebox{0.72}{
\begin{tabular}{llcccc}
\toprule
Datasets & Methods & BWT & FWT & Task Acc. (\%) & Preq. Acc. (\%) \\
\midrule
RMNIST & ADCN & -0.89 $\pm$ 1.1 & \textbf{41 $\pm$ 1.35} & \textbf{79.64 $\pm$ 0.68} & \textbf{78.41  $\pm$ 0.46} \\
 & STAM & \textbf{0.9 $\pm$ 0.35} & 30 $\pm$ 0.37$^\times$ & 77.58 $\pm$ 0.43$^\times$ & 74.71 $\pm$ 0.29$^\times$ \\
 & DCN+LwF & -15 $\pm$ 6.54$^\times$ & 16 $\pm$ 5.50$^\times$ & 39.47 $\pm$ 10.13$^\times$ & 52.79 $\pm$ 13.54$^\times$ \\
 & DCN+SI & -13 $\pm$ 6.00$^\times$ & 17 $\pm$ 6.92$^\times$ & 44.62 $\pm$ 12.46$^\times$ & 55.66 $\pm$ 14.83$^\times$ \\
 & AE+KM+LwF & -18 $\pm$ 2.32$^\times$ & 18 $\pm$ 1.79$^\times$ & 45.31 $\pm$ 1.63$^\times$ & 60.15 $\pm$ 1.54$^\times$\\
 & AE+KM+SI & -9 $\pm$ 2.72$^\times$ & 16 $\pm$ 2.15$^\times$ & 49.07 $\pm$ 0.73$^\times$ & 51.19 $\pm$ 1.39$^\times$\\
 \midrule
PMNIST & ADCN & -20 $\pm$ 2.72 & 2 $\pm$ 2.23 & 21.18 $\pm$ 1.76 & 34.88 $\pm$ 1.84 \\
 & STAM & \textbf{0.3 $\pm$ 0.09} & 1 $\pm$ 0.44$^\times$ & \textbf{47.97 $\pm$ 0.59} & \textbf{55.37 $\pm$ 0.26} \\
 & DCN+LwF & -30 $\pm$ 1.70$^\times$ & {3  $\pm$ 1.42} & {35.53  $\pm$ 0.78} & 56.50 $\pm$ 0.54 \\
 & DCN+SI & -43 $\pm$ 3.23$^\times$ & 1 $\pm$ 1.01 & 33.09  $\pm$ 2.14 & {64.87 $\pm$ 0.31} \\
 & AE+KM+LwF & -28 $\pm$ 1.71$^\times$ & {3  $\pm$ 1.85} & 35.53  $\pm$ 1.02 & 56.27 $\pm$ 0.55 \\
 & AE+KM+SI & -35 $\pm$ 2.35$^\times$ & 1 $\pm$ 1.97 & 36.06 $\pm$ 1.23 & 61.5 $\pm$ 0.36 \\
\midrule
 SMNIST & ADCN & -9 $\pm$ 1.33 & 12 $\pm$ 2.53 & {83.40 $\pm$ 1.96} & {90.96 $\pm$ 0.59} \\
 & STAM & \textbf{-2 $\pm$ 0.13} & 0$^\times$ & \textbf{92.18 $\pm$ 0.32} & \textbf{91.98 $\pm$ 0.32} \\
 & DCN+LwF & -7 $\pm$ 3.98 & 18 $\pm$ 1.17 & 52.42 $\pm$ 5.02$^\times$ & 53.46 $\pm$ 2.25$^\times$ \\
 & DCN+SI & -4 $\pm$ 1.45 & \textbf{22 $\pm$ 2.89} & 58.82 $\pm$ 1.18$^\times$ & 57.00 $\pm$ 0.34$^\times$ \\
 & AE+KM+LwF & -5 $\pm$ 1.11 & 18 $\pm$ 1.03 & 55.12 $\pm$ 0.97$^\times$ & 54.69 $\pm$ 0.58$^\times$ \\
 & AE+KM+SI & {-3 $\pm$ 1.88} & 22 $\pm$ 1.73 & 58.84 $\pm$ 0.71$^\times$ & 56.58 $\pm$ 0.28$^\times$\\
 \midrule
 SCIFAR10 & ADCN & -22 $\pm$ 0.82 & \textbf{17 $\pm$ 3.01} & \textbf{26.42 $\pm$ 0.46} & \textbf{43.96 $\pm$ 1.85} \\
 & STAM & -18 $\pm$ 2.46 & 0$^\times$ & 20.6 $\pm$ 0.66$^\times$ & 35.43 $\pm$ 1.22$^\times$ \\
 & DCN+LwF & \textbf{-13 $\pm$ 1.03} & 11 $\pm$ 1.12$^\times$ & 24.58 $\pm$ 0.39$^\times$ & 32.67 $\pm$ 0.34$^\times$ \\
 & DCN+SI & {-14 $\pm$ 0.72} & 11 $\pm$ 0.99$^\times$ & 24.36 $\pm$ 0.74$^\times$ & 32.49 $\pm$ 0.17$^\times$ \\
 & AE+KM+LwF & {-14 $\pm$ 0.79} & 11 $\pm$ 1.17$^\times$ & 24.00  $\pm$ 0.47$^\times$ & 32.63 $\pm$ 0.07$^\times$\\
 & AE+KM+SI & -14 $\pm$ 0.52 & 10 $\pm$ 0.51$^\times$ & 24.36 $\pm$ 0.66$^\times$ & 32.25 $\pm$ 0.27$^\times$ \\
\bottomrule
\end{tabular}
}
\end{center}
\end{table}



{Separately, it is observed that the performance of ADCN on SMNIST is second only to that of STAM. This is understood since STAM adopts a hierarchy of increasing receptive fields as the feature extractor. This structure is more robust against catastrophic forgetting because each receptive field is able to store a specific knowledge. On the other hand, ADCN employs neural network architectures for feature extraction. One reason neural networks are prone to catastrophic forgetting is that the knowledge of all tasks is likely to be distributed across all nodes, rather than secluded to specific groups of nodes \cite{french1999catastrophic}. Utilizing neural network structures, however, enables ADCN to perform the learning process on GPU and thus accelerates the learning process. For instance, the average execution time of ADCN on all cases to train and test a data batch is around 70 seconds, whereas STAM requires more than 240 seconds to complete both processes. This fact highlights that ADCN is more applicable for practical implementation in streaming environments.}


Also, it is noticed that ADCN's numerical results are inferior to other methods on the PMNIST problem. We also notice that the performances of other baselines are unsatisfactory on this problem. This finding is likely attributed to the abrupt-change nature of the problem that occurred between tasks. The relatively low FWT on this problem supports our argument, signifying that there is almost no relevant knowledge to improve the performance in incoming tasks. It is understood that the latent representation learned by a DNN tend to be stable or less varying. When the task changes, the update forces the parameters to move to a point optimizing the performance on the new task resulting in changing representation. The implementation of $L_{CL}$, LwF and SI try to consolidate the amount of update such that the network is capable of generating the stable latent representation. Changes that occurred on the PMNIST problem, however, are too severe to be addressed by these strategies in a single-pass learning fashion without any supervision. {Another rationale behind low results in the PMNIST problem is attributed by the absence of any labeled samples for model updates. That is, the structural evolution is only guided by the reconstruction loss meaning that it only adapts to the virtual drift.} 

The comparison results against all baselines demonstrate the efficacy of the ADCN self-organizing mechanism. Note that all consolidated algorithms possess the same network capacity at the beginning of the training process. Moreover, all baselines start with 500 clusters, whereas ADCN only consists of 2 clusters before the process runs. ADCN, however, is capable of increasing its network and learning capacity on demand. The uncovered latent representation can be immediately accommodated via a cluster growing mechanism. On the other hand, all baselines possess a static learning capacity requiring them to replace the stored knowledge in the cluster with a new one whenever there are new instances or classes. Because of this reason, their accuracy drastically suffers in an unsupervised continual learning scenario. This finding suggests that the evolving trait of ADCN took part in both delivering better predictive performance and preventing catastrophic forgetting.

\subsection{Ablation Study}
The effectiveness of the ADCN evolving mechanism in handling unsupervised problems both in standard or continual learning environments has been demonstrated in two simulation scenarios. In this subsection, we further evaluate the effectiveness of latent-based regularization $L_{CL}$ in handling catastrophic forgetting. Also, we evaluate the effect of reducing the number of labeled samples for associating centroids with classes $N_m$. The first ablation is only conducted on RMNIST and SMNIST problems attempting to obtain obvious findings, whereas the second ablation also involves MNIST and FMNIST datasets.
\begin{table}[t!]
\caption{Performance Metrics on Ablation Study.}
\label{result-ablation}
\begin{center}
\scalebox{0.85}{
\begin{tabular}{llrrr}
\toprule 
 & Ablations & BWT & FWT & Task Acc. (\%)\tabularnewline
\midrule 
RMNIST & ADCN & \textbf{-0.89 $\pm$ 1.1} & 41 $\pm$ 1.35 & \textbf{79.64 $\pm$ 0.68}\tabularnewline
 & ADCN w/o $L_{CL}$ & -13 $\pm$ 1.18$^{\times}$ & \textbf{43 $\pm$ 1.04} & 70.86 $\pm$ 1.05$^{\times}$\tabularnewline
\midrule 
SMNIST & ADCN & \textbf{-9 $\pm$ 1.33} & 12 $\pm$ 2.53 & \textbf{83.40 $\pm$ 1.96}\tabularnewline
 & ADCN w/o $L_{CL}$ & -10 $\pm$ 3.49 & \textbf{15 $\pm$ 2.35} & 72.78 $\pm$ 5.16$^{\times}$\tabularnewline
\bottomrule
\end{tabular}
}
\end{center}
\end{table}

From Table \ref{result-ablation}, it is clear that removing $L_{CL}$ dramatically decreases both task accuracy and BWT indicating that the respected ablation suffers from a catastrophic forgetting problem. It is understood since the implementation of $L_{CL}$ stabilizes the latent representation output and thus creating a task-invariant network. Even the previous task representations have been stored in the centroids as patterns, yet none of them are recognized by the clusters without implementing $L_{CL}$ since the previous task representations are changing after learning a new task. This suggests that the implementation of $L_{CL}$ completes the ADCN learning policy in handling continual unsupervised learning problems. Another interesting observation is that the unregularized ADCN delivered better FWT. This is explainable since the network parameters having a good initial condition can freely move to the cost-optimal point of the new task. Also, the stored knowledge in the clusters helps to achieve better accuracy in the incoming tasks. This supports the implementation of the ADCN learning policy to handle transfer learning problems in an unsupervised manner.

\begin{figure}[!t]
\centerline{\includegraphics[scale=0.6]{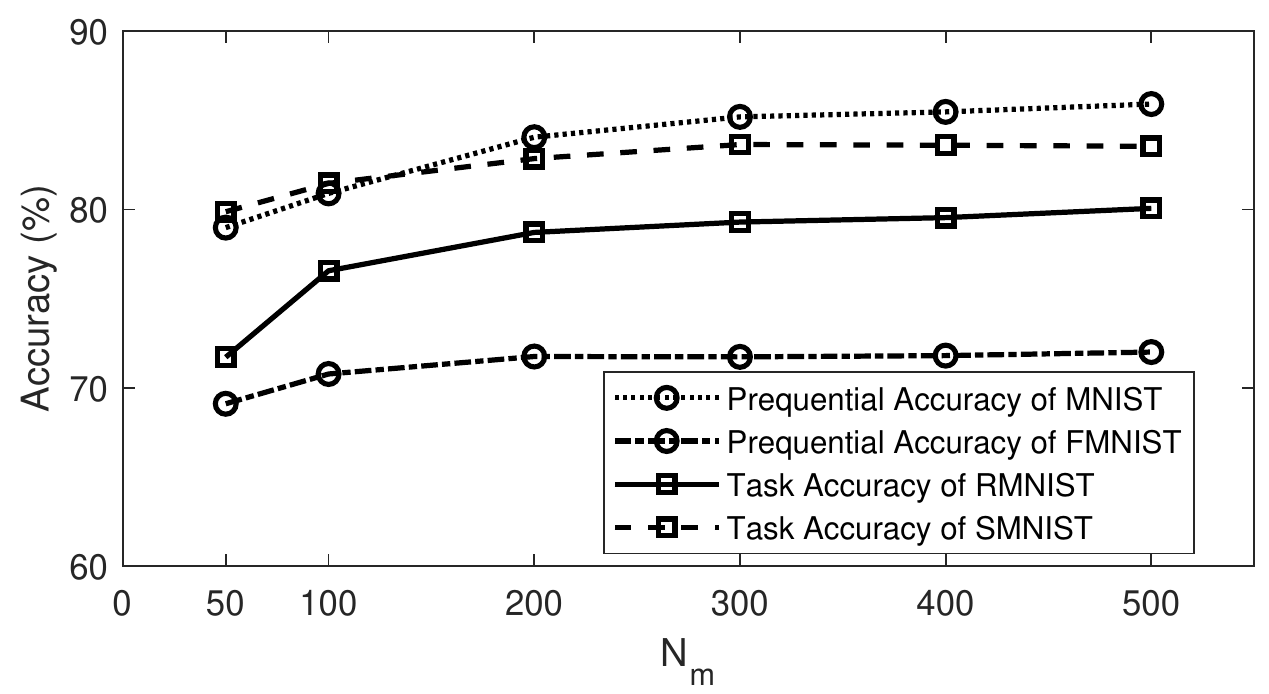}}
\caption{The number of labeled samples per class $N_m$ for associating centroids with classes is varied.}
\label{fig:scenario2}
\end{figure}

Separately, the second ablation results are depicted in Fig. \ref{fig:scenario2}. We consider $N_m = \{50,100,200,300,400,500\}$. It is observed that the performance of ADCN does not decrease significantly when reducing $N_m$ to 200. The task accuracy on RMNIST and SMNIST problems also indicates that ADCN experiences less catastrophic forgetting. It is understood since the learning process is only dependent on streaming unlabeled samples. The SAE and self-clustering mechanism are capable of updating themselves in an unsupervised manner. However, we observe that ADCN delivered a comparable performance with fewer labeled samples compared to several baselines enjoying $N_m=500$. In terms of task accuracy, for reference, ADCN achieved  $79.87\%$ ($N_m=50$), whereas DCN+SI attained $58.82\%$ ($N_m=500$) on SMNIST problem.

\subsection{Theoretical Study}
{This section is meant to study the validity of ADCN's structural learning in which node, layer and cluster are grown and pruned dynamically from streaming data. Our study is based on the statistical learning theory analyzing the generalization bound of a model \cite{vapnik2013nature}. The generalization bound of a model is defined in term of the expected risk function $R_{exp}(f)$  having an upper bound as a factor of the empirical risk function $R_{emp}(f)$ plus its variance $\sigma_{emp}^2$. It is formally introduced:}
\begin{equation}\label{upperbound}
    R_{exp}(f) \leq R_{emp}(f)+\sigma_{emp}^2 
\end{equation}
{where $R_{emp}(f)$ is calculated in term of prediction error.}

\begin{figure}[!t]
\centerline{\includegraphics[scale=0.33]{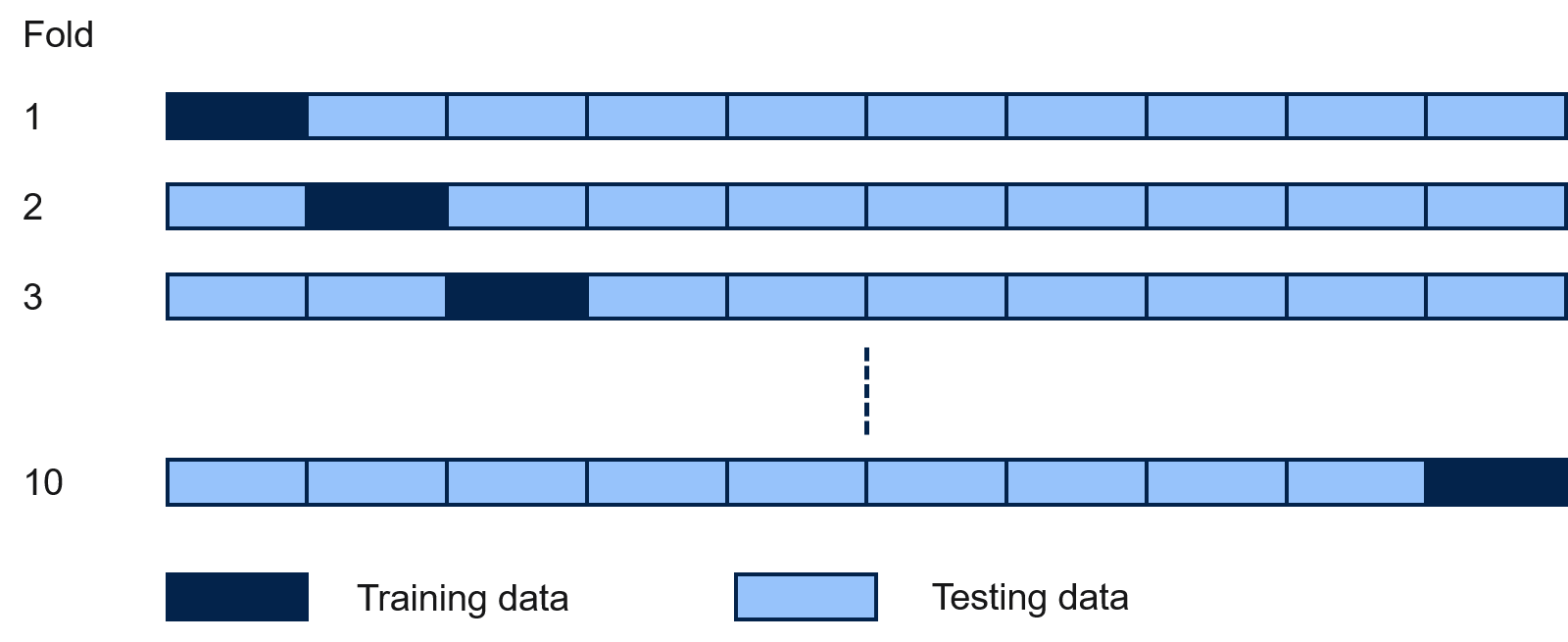}}
\caption{The ten fold reversed cross validation.}
\label{fig:reversedCV}
\end{figure}

{Direct evaluation of the expected risk function $R_{exp}(f)$ is impractical since it calls for the presence of unseen samples. Hence, the only way for the evaluation of generalization power is via the upper bound of (\ref{upperbound}) which can be estimated using the reversed cross validation procedure. The $k$-fold reversed cross validation protocol distinguishes itself from the conventional cross validation in the proportion of training samples where only one data bin or $1/k^{th}$ part of data samples are made available in the training process while leaving the remainder of data samples in the testing process. That is, all data samples are split into $k$ equal-sized data bins. Fig. \ref{fig:reversedCV} visualizes the $5$-fold reversed cross validation. The reversed cross validation protocol portrays an upper bound of model's generalization power because it presents a challenging training condition where only few training samples are accessible. The upper bound of generalization power is ultimately recorded as $R_{emp}(f)+\sigma_{emp}^2$ where $R_{emp}(f)$ depicts the testing classification loss measured in every fold of the reversed cross validation mechanism while $\sigma_{emp}^{2}$ is its variance after visiting the testing process of all $k$ data bins.} 

{Since the focus of this study is to assess the quality of network structure evolved during the training process, ADCN's parameters after the training process are randomly reinitialized and retrained from scratch with the absence of structural learning phase. Note that the incremental clustering mechanism is still carried out in this case because random initialization of cluster centroids leads to loss of generalization power. That is, cluster's centroid no longer represents original data points. Our experiment is carried out in the MNIST dataset consisting of $70K$ data samples. This implies the training set size is $7K$ while the testing set involves $63K$ data samples. ADCN is compared with STAM \cite{smith2019unsupervised} also being a unsupervised continual learning and using a handcrafted network structure.}  
\begin{table}[t!]
\caption{Network structure performance on MNIST problem.}
\begin{centering}
\label{gentestmnist} \scalebox{1}{ %
\begin{tabular}{lrr}
\toprule 
Fold & \multicolumn{2}{c}{$R_{emp}(f)$}\tabularnewline
 & ADCN & STAM\tabularnewline
\midrule 
1 & 25.01 & 28.87 \tabularnewline
2 & 25.60 & 29.25 \tabularnewline
3 & 29.50 & 29.64 \tabularnewline
4 & 24.82 & 27.28 \tabularnewline
5 & 24.94 & 26.08 \tabularnewline
6 & 27.12 & 24.86 \tabularnewline
7 & 26.22 & 24.05 \tabularnewline
8 & 26.36 & 30.32 \tabularnewline
9 & 26.55 & 29.71 \tabularnewline
10 & 25.06 & 27.47 \tabularnewline
\midrule
$\sigma_{emp}^2$ & 1.84 & 4.27 \tabularnewline
$\max(R_{emp}(f))$ & 29.5 & 30.32 \tabularnewline
$\max(R_{emp}(f)) + \sigma_{emp}^2$ & \textbf{31.34} & 34.59 \tabularnewline
\bottomrule
\end{tabular}
} 
\par\end{centering}
\vspace{5pt}
\end{table}

{Table \ref{gentestmnist} reports the error rate of ADCN and STAM where the reversed cross validation is undertaken across 10 folds. It is presented that ADCN generated network attained the lowest upper-bound value of error rate ($\max(R_{emp}(f)) + \sigma_{emp}^2$ = 31.34) which indicates that the performance of ADCN generated network is better than STAM network on unseen samples possessing the same data distribution. The generalization performance test also validates the ADCN performance on MNIST reported in \textit{Table \ref{unsupervisedlearningresults}} and at the same time confirms the effectiveness of ADCN learning policy in generating a network structure for any given problem in an unsupervised manner.}

\subsection{Related Works and Discussion}
Even though deep clustering network has been successfully used in the past to solve unsupervised learning problem \cite{liu2005toward,Caron_2018_ECCV,ge2019dual}, yet most of them are incapable of expanding their learning capacity and have not been tested to process streaming data in the continual learning environment. ADCN, on the other hand, is designed to take advantage of both the self-evolving mechanism and latent-based regularization to handle such situations. The SAE evolution aims to adapt to any concept changes increasing the network capacity, whereas the cluster growing mechanism expands ADCN knowledge covering new latent representation. The implementation of latent-based regularization completes the learning policy as an effort to mitigate the catastrophic forgetting issue.

DeepCluster is introduced in \cite{Caron_2018_ECCV} to jointly learn the parameters of DNNs and clusters. It is also capable of generating pseudo-labels to update the network in a supervised manner. Dual-AAE is proposed in \cite{ge2019dual} to distill classification information from the unlabeled data. It involves adversarial training in the learning process to obtain better latent representations. MIX'EM is proposed in \cite{varamesh2020mix} as a novel solution of unsupervised image classification. By constructing a mixture
of embedding module, it is capable of producing sufficient representations for driving the clustering process. In this study, however, we do not compare the performance of ADCN with DeepCluster, Dual-AAE and MIX'EM because they highly depend on the iterative training process to achieve the best performance hindering their implementation in streaming environments. Our method, on the other hand, is applicable to such cases and is reliably capable of constructing a good quality network and clusters even only utilizing a GPU GeForce GTX 1080 in just 5-10 minutes.

ParsNet \cite{parsnet} is proposed to handle the lack of labeled samples in streaming environments via network evolving mechanism. CBLN utilizes Bayesian NN to flexibly allocate additional learning resources to adapt to new tasks for mitigating the catastrophic forgetting problem \cite{CLBN}. Nonetheless, those methods involve a supervised learning mechanism requiring labeled samples before the process runs which does not fit with our simulation scenario. STAM, a non-DNN based approach \cite{smith2019unsupervised}, puts forward an online clustering mechanism for unsupervised continual learning in non-stationary environments. It is capable of expanding the learning capacity attempting to prevent catastrophic forgetting. STAM, however, does not benefit from the neural network structure \cite{liu2005toward,Caron_2018_ECCV}. {Employing uncommon feature extractor causes STAM to have lack of support from the most library in deep learning community for executing both training and testing process on GPU.} Table S-III of the supplemental document summarizes the characteristics of the aforementioned methods.

We agree that to create an algorithm possessing human-level intelligence we should follow the biological learning philosophy. That is, the designed algorithm should possess the ability to handle any concept changes or any incoming task without any labels in a single-pass learning fashion. Our proposed method, ADCN, pursues the same direction as the philosophy. It is capable of continually learning from the streaming unlabeled data. When a classification task is required, one surely needs to reveal several labeled samples for associating centroids with classes. It is worth mentioning that ADCN does not use these labels to perform a training process. The learning capacity of ADCN can be incrementally increased via network and cluster evolution. The network structure can be flexibly adjusted in respect to problem complexity, whereas the clusters can be constructed autonomously accommodating newly seen knowledge. Even though ADCN has no access to previously seen data, it is still capable of mitigating the catastrophic forgetting problem. This is explainable since the previously seen knowledge has been stored as centroids. Further, ADCN stabilizes the latent representation via latent-based regularization attempting to obtain task-invariant network. Interestingly, our finding also signifies that ADCN is applicable to handle a transfer learning problem in streaming environments. That is, the crafted knowledge in the clusters can help to improve the predictive performance of the incoming tasks.

\section{Conclusion}
This brief introduces a novel Autonomous Deep Clustering Network (ADCN). It simultaneously optimizes the deep network and cluster parameters via a joint optimization of reconstruction loss and clustering loss. The learning capacity is expandable via incremental learning of SAE and self-clustering mechanisms. ADCN also incorporates a latent-based regularization to further mitigate the catastrophic forgetting problem. We evaluate our proposed method on standard unsupervised learning and unsupervised continual learning scenarios. The results show that it can outperform other baselines in almost all cases. Since ADCN exhibits a promising positive forward transfer performance, we plan to extend this algorithm to unsupervised transfer learning in the future. Also, it would be intriguing to derive sufficient conditions to guarantee the performance of the ADCN generated network.

\ifCLASSOPTIONcaptionsoff
  \newpage
\fi





%

\bibliographystyle{IEEEtran}
\bibliography{mbibfile}




%








\end{document}


\title{Supplementary Material of Unsupervised Continual Learning in Streaming Environments}


\author{Andri Ashfahani and Mahardhika~Pratama,~\IEEEmembership{Senior Member,~IEEE}
} 

\maketitle

\begin{abstract}
This document provides supplementary material for our paper titled Unsupervised Continual Learning in Streaming Environments. It consists of the list of symbols and acronyms, ADCN learning algorithm, the illustration of two simulation scenarios and a figure of performance and network evolution. To support the reproducible research initiative, codes and raw results of ADCN are made available in \url{https://tinyurl.com/AutonomousDCN}.
\end{abstract}

\IEEEpeerreviewmaketitle

\section{Symbols and Acronyms}
To facilitate the reader in understanding our paper, we provide a list of symbols and acronyms used in the paper. It is presented in Table \ref{symbols} and Table \ref{acronyms}, respectively.
\begin{table}
\renewcommand\thetable{S-I}
\caption{Symbols}
\label{symbols}
\begin{center}
\scalebox{1}{
\begin{tabular}{ll}
\toprule 
\multicolumn{2}{c}{Symbols}\tabularnewline
\midrule 
$B_{k}$ & the $k-th$ data batch\tabularnewline
$X_{k}$ & the input data in the $k-th$ batch\tabularnewline
$X_{t}$ & the data point at the $t-th$ time stamp\tabularnewline
$N$ & the number of data in a batch\tabularnewline
$N_{init}$ & the number of initial data for pretraining phase\tabularnewline
$N_{m}$ & the number of labeled data per class\tabularnewline
 & for asscociating centroids with classes \tabularnewline
$nE$ & the number of epoch in the pretraining phase\tabularnewline
$nT$ & the number of tasks\tabularnewline
$K$ & the number of bathces\tabularnewline
$u$ & the size of input $X$\tabularnewline
$M$ & the number of classes\tabularnewline
$M_{iT}$ & the number of classes at the $iT-th$ task\tabularnewline
$F(.)$ & the feature extractor\tabularnewline
$Z$ & the output of the feature extractor\tabularnewline
$h^{l}$ & the output of the $l-th$ SAE layer\tabularnewline
$h^{l}_n$ & $h^{l}$ generated by labeled samples\tabularnewline
$W^{l}$ & the weight of the $l-th$ SAE layer\tabularnewline
$b^{l}$ & the encoder bias of the $l-th$ SAE layer\tabularnewline
$d^{l}$ & the decoder bias of the $l-th$ SAE layer\tabularnewline
$u_{l}$ & the number of inputs of the $l-th$ SAE layer\tabularnewline
$R_{l}$ & the number of nodes of the $l-th$ SAE layer\tabularnewline
$L$ & the number of SAE layers \tabularnewline
$c_{j}^{l}$ & $j-th$ centroid at the $l-th$ cluster\tabularnewline
$C_{l}$ & the number of cluster at the $l-th$ SAE layer\tabularnewline
$Sup_{j}^{l}$ & the support of $c_{j}^{l}$ cluster\tabularnewline
$a_{c_{j},h_{n}}^{l}$ & the $c_{j}^{l}$ cluster's allegiance given $h^l_{n}$\tabularnewline
$N_{m}$ & the number of labeled data per class\tabularnewline
$A_{c_{j},m}^{l}$ & the allegiance of $c_{j}^{l}$ cluster to the $m-th$ class\tabularnewline
$score^{l}$ & the prediction score from clusters at $l-th$ SAE layer\tabularnewline
$\hat{Y}$ & the predicted label\tabularnewline
$\mu_{a}^{t}$ & the recursive mean of array $a$\tabularnewline
$\sigma_{a}^{t}$ & the recursive standard deviation of array $a$\tabularnewline
$\mu_{a}^{min}$ & the minimum value of $\mu_{a}^{t}$\tabularnewline
$\sigma_{a}^{min}$ & the minimum value of $\sigma_{a}^{t}$\tabularnewline
$D(x,y)$ & the L2 distance of vector $x$ and $y$\tabularnewline
$k^l_{win}$ & the dynamic confidence degree of \tabularnewline
 & cluster growing mechanism\tabularnewline
$\alpha_{x,w,d}$ & the significance level of drift detector\tabularnewline
$\epsilon$ & the error bound\tabularnewline
$S$ & the output of feature extractor $F(.)$ given $[B_{k-1};B_{k}]$\tabularnewline
$\alpha$ & the clustering loss strength\tabularnewline
$L_{U}$ & the unsupervised learning objective\tabularnewline
$L_{CL}$ & the latent-based regularization\tabularnewline
$L_{UCL}$ & the unsupervised continual learning objective\tabularnewline
$\lambda_{iT}$ & the latent-based regularization strength for $iT-th$ task\tabularnewline
$\beta$ & the maximum value of $\lambda_{iT}$\tabularnewline
\bottomrule
\end{tabular}
}
\end{center}
\end{table}

\begin{table}
\renewcommand\thetable{S-II}
\caption{Acronyms}
\label{acronyms}
\begin{center}
\scalebox{1}{
\begin{tabular}{ll}
\toprule 
\multicolumn{2}{c}{Acronyms}\tabularnewline
\midrule 
ADCN & Autonomous Deep Clustering Network\tabularnewline
DNN & Deep Neural Network\tabularnewline
ODL & Online Deep Learning\tabularnewline
ADL & Autonomous Deep Learning\tabularnewline
MLP & Multi-layer Perceptron\tabularnewline
NADINE & Neural Network with Dynamic Capacity\tabularnewline
PALM & Parsimonious Learning Machine\tabularnewline
COMPOSE & Compacted Object Sample Extraction\tabularnewline
SCARGC & Stream Classification Algorithm Guided by Clustering\tabularnewline
SLASH & Self-labeling with Hedge\tabularnewline
NS & Network Significance\tabularnewline
CNN & Convolutional Neural Network\tabularnewline
SAE & Stacked Autoencoder\tabularnewline
AE & Autoencoder\tabularnewline
ReLU & Rectified Linear Unit\tabularnewline
MSE & Mean Squared Error\tabularnewline
SGD & Stochastic Gradient Descent\tabularnewline
SVM & Support Vector Machine\tabularnewline
KNN & K-Nearest Neighbour\tabularnewline
Preq. Acc. & Prequential Accuracy\tabularnewline
LwF & Learning without Forgetting\tabularnewline
SI & Synaptic Intelligence\tabularnewline
DCN & Deep Clustering Network\tabularnewline
KM & K-Means\tabularnewline
AAE & Adversarial Autoencoder\tabularnewline
CBLN & Continual Bayesian Learning Network\tabularnewline
STAM & Self-Taught Associative Memory\tabularnewline
\bottomrule
\end{tabular}
}
\end{center}
\end{table}

\section{ADCN Training Algorithm}
Algorithm \ref{algo:adcnLP} presents ADCN learning mechanism in handling a batch of data $X$. It is obvious that it consists of two main training mechanism, i.e, the optimization of parameters and structure of the network and the cluster. It is conducted alternately for solving a joint optimization problem. In the case of unsupervised learning (UL), the loss function in Equation (14) is employed to update ADCN parameters, whereas Equation (16) is utilized to adjust the parameters in unsupervised continual learning (UCL) situation. In the pretraining phase, multiple epochs $nE=50$ is used, whereas ADCN applies one-pass learning fashion $nE=1$ in the streaming phase attempting to cope with the rapid nature of data streams.
\begin{algorithm}
\small
	\caption{ADCN learning policy.}
	\label{algo:adcnLP}
	     
		\textbf{Require:} network, $X$, $\{\alpha_{x},\alpha_{d},\alpha_{w}\}$,  $\alpha$, $\beta$ and $nE$

		\textbf{Ensure:}  $Scenario \in \{UL, UCL\}$
		
		/*The adjustment of feature extractor and the incremental learning of SAE*/
		
		\For {$iE = 1$ to $nE$}{
		    
		    \textbf{Execute:} Network depth adaptation via Equation (10)-(12)
		
    		\For {$n = 1$ to $N$}{
    		    \textbf{Execute:} Network width adaptation via Equation (7)-(9)
    		    
    		    \eIf{Scenario is UL}{
    			 \textbf{Update:} Network parameters by minimizing the loss function in Equation (14)
    			 }
    	    	 {
    	    	 \textbf{Update:} Network parameters by minimizing the loss function in Equation (16)
    	    	 }	
    		}
		    
		}
		
		/*Self-clustering mechanism*/
		
		\For {$iE = 1$ to $nE$}{
		    
		    \For {$n = 1$ to $N$}{
    		    \textbf{Find:} The winning cluster via $win\rightarrow \min_{i=1,...,C_l}D(h^l,c^l_{win})$
    		    
    		    \eIf{ Equation (13) is satisfied }
    		    {
    		    
    		    \textbf{Execute:} Cluster growing
    		    
    		    }
    		    {
    		    
    		    \textbf{Update:} The winning cluster via Equation (15)
    		    
    		    }
    		}
		    
		}

		\textbf{Return:} network
\end{algorithm}

\section{Simulation Protocols}
\begin{figure}
\renewcommand{\thefigure}{S-1}
\centerline{\includegraphics[scale=0.43]{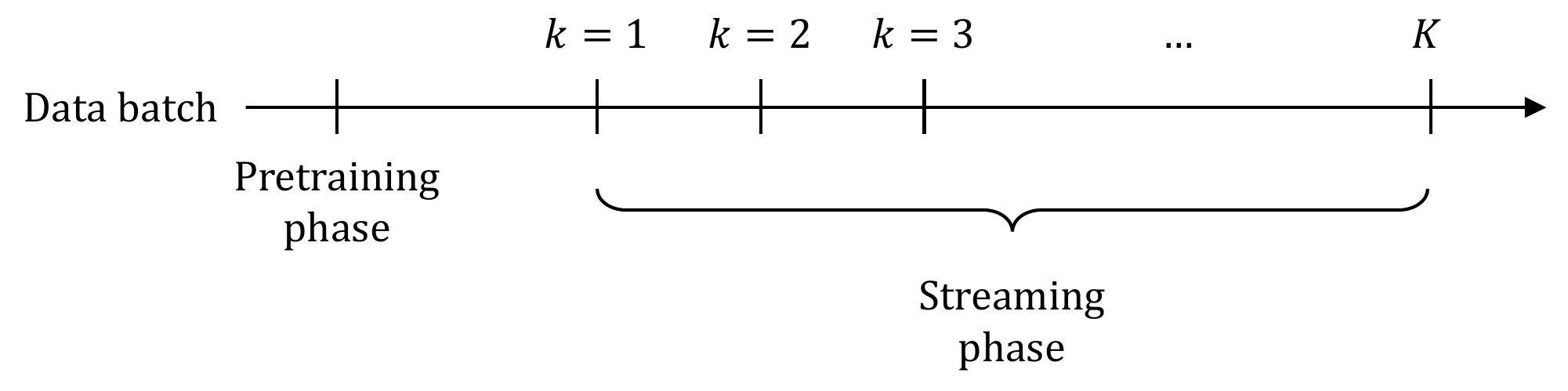}}
\caption{Unsupervised learning simulation scenario. The algorithm performance in dealing with concept drift is tested here.}
\label{fig:ul}
\end{figure}

The simulation scenario of the first experiment is depicted in Fig. \ref{fig:ul}. Since the prequential procedure is adopted, ADCN performs training and testing processes in every data batch. One may reveal several labeled samples per class in every batch. In our simulation scenario, however, $N_m$ labeled samples are only revealed in the first batch $k=1$ simulating a real-world and the most difficult scenario. ADCN is required to autonomously adapt to any changes from the streaming unlabeled samples. The prequential accuracy of ADCN can be calculated by averaging the accuracy of predictions across all batches.

\begin{figure*}[!t]
\renewcommand{\thefigure}{S-2}
\centerline{\includegraphics[scale=0.43]{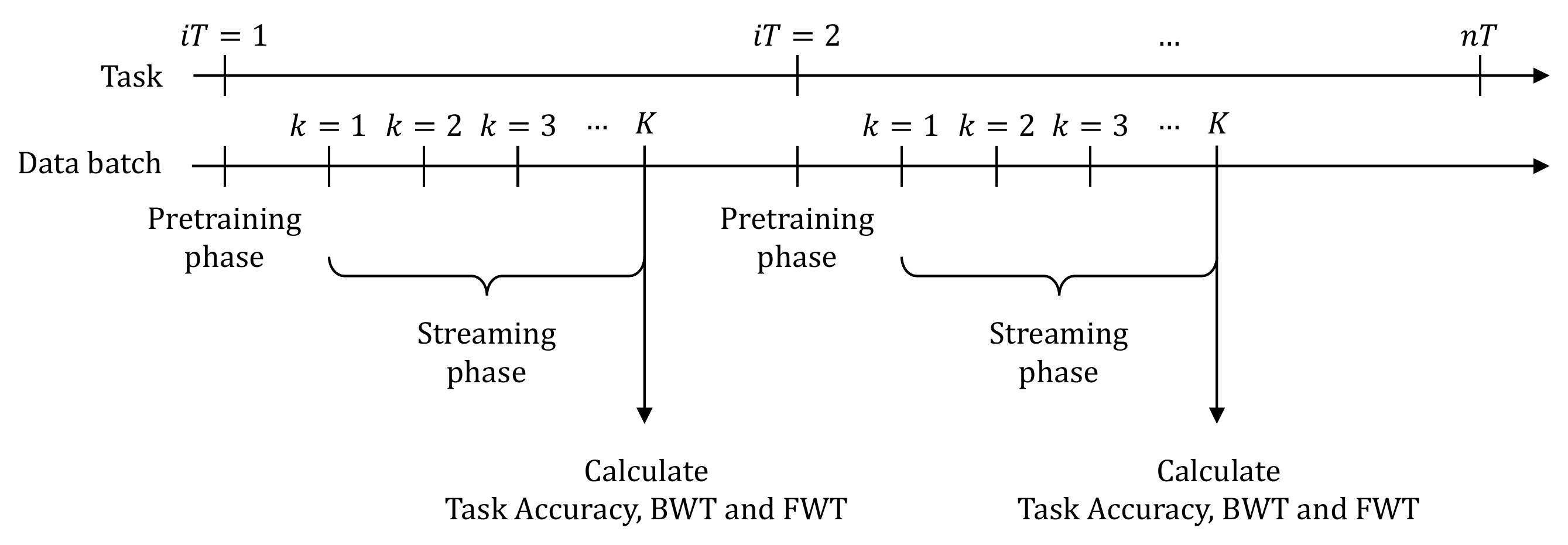}}
\caption{Unsupervised continual learning simulation scenario. It simulates a streaming environment consisting of more than a task ($nt>1$). It tests an algorithm's capability in adapting to concept changes and in resolving the catastrophic forgetting problems.}
\label{fig:ucl}
\end{figure*}

Fig. \ref{fig:ucl} illustrates the simulation scenario of an unsupervised continual learning experiment. In this scenario, ADCN is required to process more than one task ($nT>1$). Note that it has no access to the unlabeled data stream of the previous task. $N_{init}$ and $N_m$ data are evenly divided into $nT$ tasks leading to ($N_{init}/nT$) and ($N_m/nT$) data per task. A small set of labeled data is revealed every $k=1$ in each task for associating centroids with classes. Besides prequential accuracy, we also calculate task accuracy, BWT and FWT which take place in the last batch of each task. That is, the capability of ADCN in both mitigating catastrophic forgetting and exhibiting positive knowledge transfer are tested here. We exclusively provide a batch of unseen data per task for calculating those three metrics.

From Fig. \ref{fig:ucl}, it is obvious that we assume the task boundary is known in the training process. However, ADCN is capable of predicting any samples from previously seen tasks without knowing the task ID in the testing process. As many as $N_m$ labeled samples surely need to be given for associating centroids with classes. This practice is still inline with biological learning philosophy. For example, a student is fully aware that he has finished studying a subject and is going to study another subject. In the exam, however, he can freely choose which answer should be provided for a given question. Note that all baselines are implemented in the same simulation scenario.

\section{Performance and Network Evolution}
\begin{figure}[!t]
\renewcommand{\thefigure}{S-3}
\centerline{\includegraphics[scale=0.5]{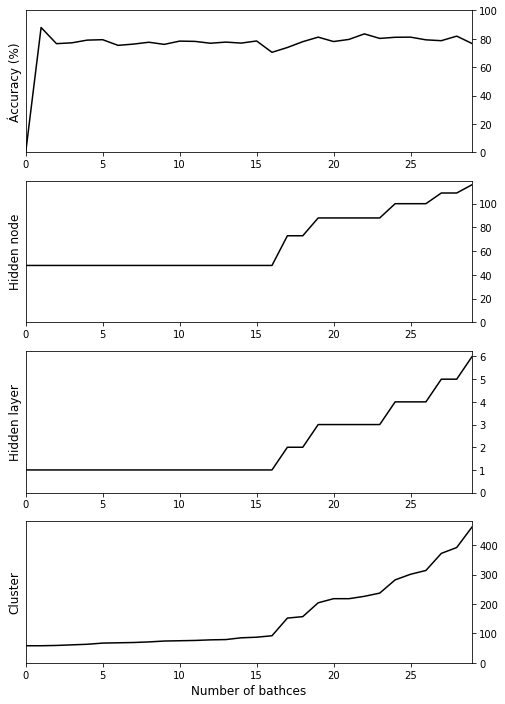}}
\caption{Performance of ADCN on Default of Credit Card Clients.}
\label{fig:creditevo}
\end{figure}

Fig. \ref{fig:creditevo} presents the performance and network evolution on Credit Card Default problem. It demonstrates the efficacy of the ADCN learning policy. The accuracy increases as the number of samples increases. The learning capacity of ADCN is expandable following the problem complexity. The capacity is timely increased in the case of drift indicated by a small decrease in accuracy. All in all, ADCN is capable of generating a proper network complexity for a given problem in an unsupervised learning situation.

\section{prominent Characteristics of ADCN and Its Related Works}
Table \ref{prominentchar} presents the prominent characteristics of consolidated related works and ADCN. It is obvious that ADCN characterizes flexible learning capacity which addresses unsupervised and continual learning problems. It is executable in a streaming environment. Further, built upon DNN structure enables ADCN to process wider variety of data streams.
\begin{table*}
\renewcommand\thetable{S-III}
\caption{Prominent Characteristics of ADCN and Its Counterparts}
\label{prominentchar}
\begin{center}
\scalebox{1}{
\begin{tabular}{lrrrrrrr}
\toprule 
Characteristics & ADCN & DeepCluster \cite{Caron_2018_ECCV} & Dual-AAE \cite{ge2019dual} & MIX'EM \cite{varamesh2020mix} &ParsNet \cite{parsnet} & CBLN \cite{CLBN} & STAM \cite{smith2019unsupervised} \tabularnewline
\midrule 
Online learning & $\checkmark$ &  &  &  & $\checkmark$ &  & $\checkmark$ \tabularnewline
DNN-based approach & $\checkmark$ & $\checkmark$ & $\checkmark$ & $\checkmark$ & $\checkmark$ & $\checkmark$ & \tabularnewline
Unsupervised learning & $\checkmark$ & $\checkmark$ & $\checkmark$ & $\checkmark$ &  &  & $\checkmark$\tabularnewline
Dynamic learning capacity & $\checkmark$ &  &  &  & $\checkmark$ & $\checkmark$ & $\checkmark$ \tabularnewline
Resistance to catastrophic forgetting & $\checkmark$ &  &  &  &  & $\checkmark$ & $\checkmark$\tabularnewline
\bottomrule
\end{tabular}
}
\end{center}
\end{table*}

\ifCLASSOPTIONcaptionsoff
  \newpage
\fi





%
\newpage
\bibliographystyle{IEEEtran}
\bibliography{mbibfile}




%







